\documentclass[10pt, a4paper]{article}

\usepackage[final]{lrec2026} %

\usepackage{todonotes}
\usepackage{amsmath}
\usepackage{amsthm}
\usepackage{bm}
\usepackage{listings}

\usepackage{algorithm}
\usepackage{algpseudocode}

\usepackage{booktabs}

\usepackage{tcolorbox}

\title{sebis at ArchEHR-QA 2026: How Much Can You Do Locally? Evaluating Grounded EHR QA on a Single Notebook}

\name{Ibrahim Ebrar Yurt\textsuperscript{1}\textsuperscript{*}, Fabian Karl\textsuperscript{1}\textsuperscript{*}, Tejaswi Choppa\textsuperscript{1}, Florian Matthes\textsuperscript{1}} 

\address{ \textsuperscript{1}Technical University of Munich,
 School of Computation, Information and Technology,
 Germany
\\
 \small{
  \{ibrahim.yurt, fabian.karl, tejaswi.choppa, matthes\}@tum.de
 }\\
     \textsuperscript{*}Shared first authorship
    }

\abstract{
Clinical question answering over electronic health records (EHRs) can help clinicians and patients access relevant medical information more efficiently. However, many recent approaches rely on large cloud-based models, which are difficult to deploy in clinical environments due to privacy constraints and computational requirements.
In this work, we investigate how far grounded EHR question answering can be pushed when restricted to a single notebook.
We participate in all four subtasks of the ArchEHR-QA 2026 shared task and evaluate several approaches designed to run on commodity hardware. All experiments are conducted locally without external APIs or cloud infrastructure.
Our results show that such systems can achieve competitive performance on the shared task leaderboards. In particular, our submissions perform above average in two subtasks, and we observe that smaller models can approach the performance of much larger systems when properly configured. These findings suggest that privacy-preserving EHR QA systems running fully locally are feasible with current models and commodity hardware.
 \\ The source code is available at \url{https://github.com/ibrahimey/ArchEHR-QA-2026}.
 \\ \newline  }

\begin{document}

\maketitleabstract
\section{Introduction}
\label{sec:introduction}

Electronic Health Records (EHRs) contain extensive clinical information about patients, including physician notes, laboratory results, medication histories, and diagnostic reports. 
Efficient access to this information is critical for both clinicians and patients. 
Question Answering (QA) over EHRs aims to provide natural language interfaces that retrieve clinically relevant information directly from medical records, thereby supporting clinical decision making and reducing physician workload~\cite{survey_ehrqa, emrQA}. 

In clinical practice, both physicians and patients frequently ask about patient histories, treatments, medications, and diagnostic results~\cite{emrQA}. 
Automating responses to such questions can improve efficiency and reduce cognitive load on healthcare professionals. 
Importantly, QA systems that ground their answers in explicit evidence from medical records can further improve transparency and trust in automated systems, especially in high-stakes domains such as healthcare.

Despite recent advances in large language models (LLMs), applying them to EHR QA presents major practical challenges. 
Clinical records contain highly sensitive personal health information and are subject to strict privacy regulations such as HIPAA and GDPR. 
Thus, medical institutions are often unable to send EHR data to external cloud services for processing~\cite{jonnagaddala2025privacy}. 
Meanwhile, modern LLM-based QA systems typically rely on large models that require specialized hardware accelerators or cloud-based inference.

In practice, this creates a gap between research and deployable clinical systems.
Many healthcare environments lack the infrastructure that is needed to host large models~\cite{oke2025infrastructure}. 
Instead, physicians often have to rely on standard workstation or notebook hardware. 
For real-world adoption, EHR QA systems must therefore be able to run entirely on local devices while still delivering acceptable performance.
 
While extensive research has focused on improving language models for biomedical and clinical tasks~\cite{singhal2023large, singhal2025toward}, relatively little work has investigated how such methods perform under strict local deployment constraints~\cite{survey_ehrqa, blavskovic2025robust}. 
Most state-of-the-art systems rely on large computational resources or cloud-hosted APIs. 
Consequently, there is a research gap in understanding which architectures and strategies remain effective even when all components must be run on a single local device.

In this work, we investigate how far grounded EHR question answering can be pushed using only locally executable models. 
We participate in all subtasks of the ArchEHR-QA~\cite{soni-etal-2026-archehr-qa} shared task and evaluate several strategies designed to operate on commodity hardware. 
Our experiments explore three classes of approaches: (1) fine-tuned BERT-style classifiers, (2) embedding-based methods, and (3) small or quantized language models for answer generation.
All experiments, including training, inference, and evaluation, are performed locally on commodity hardware without relying on external APIs or cloud infrastructure.

Our key contributions are:

\begin{itemize}
    \item \textbf{EHR QA on Commodity Hardware:} We demonstrate that a complete clinical QA pipeline can run entirely on standard commodity hardware.
    \item \textbf{Strong Embedding Baselines:} We show that out-of-the-box dense embedding models surprisingly beat fine-tuned cross-encoders for evidence extraction and alignment.
    \item \textbf{Synthetic Data for Encoder Tuning:} We highlight the brittleness of BERT-style classifiers in low-resource clinical settings and propose a local LLM-based synthetic data pipeline to help stabilize their training.
    \item \textbf{Effectiveness of Quantized LMs:} We establish that quantized and small language models are highly capable of handling the generative question interpretation and answer formulation tasks locally.
\end{itemize}

\section{Related Work}

\subsection{EHR Question Answering}

EHR question-answering systems have largely been developed and evaluated under the assumption of abundant computational resources, leaving the question of local, privacy-preserving deployment underexplored.
Early EHR QA systems were largely framed as reading comprehension and extraction tasks, where systems were asked to identify answer spans in the clinical notes \citep{survey_ehrqa}. This formulation was especially prominent in work built on emrQA \citep{emrQA}, which established span extraction as a common evaluation setup for clinical question answering and used a DrQA-style \citep{chen-etal-2017-reading} recurrent document reader as a baseline. Subsequent work moved from the earlier recurrent neural network based document readers towards transformer-based extractive QA models. In particular, \citet{soni-roberts-2020-evaluation} evaluated BERT \citep{bert}, BioBERT\citep{biobert}, and ClinicalBERT \citep{clinicalbert} on emrQA in a machine reading comprehension task. They found that intermediate fine-tuning on SQuAD \citep{squad} improved answer-span prediction and that ClinicalBERT achieved the strongest emrQA result under sequential fine-tuning. Further, \citet{lanz-pecina-2024-paragraph} proposed a two-step retrieve-then-read pipeline where the long clinical records were first segmented into paragraphs, and a retrieval model selected the most relevant segment. A QA model then extracted the answer from the retrieved context. However, \citet{PhysioNet-ehr-notes-qa-llms-1.0.0} argued that the realistic clinical QA was more complex than the earlier extractive benchmarks suggested. They noted that prior datasets largely framed EHR QA as span extraction from a single note, whereas real patient-specific questions might require synthesizing information across multiple clinical notes.
To address this, EHRNoteQA \cite{PhysioNet-ehr-notes-qa-llms-1.0.0} introduced patient-specific questions spanning ten topics, including questions that require information from multiple discharge summaries. With the rise of large language models, generative approaches also became more prominent, and they evaluated 27 LLMs in both open-ended and multiple-choice settings. The ArchEHR-QA shared task \citep{soni-demner-fushman-2026-dataset} pushed this further by requiring answers to be explicitly grounded in clinical evidence, with each answer sentence accompanied by sentence-level citations to the relevant sentences of the clinical note.

Despite this progress, the current LLM-based approaches still remain difficult to deploy in real clinical settings due to the computational costs and privacy concerns.

\subsection{Local and privacy preserving Clinical NLP}

Some of the recent work demonstrates that the privacy-preserving local deployment for clinical NLP is feasible in practice. 
\citet{Griot} implemented a GDPR-compliant LLM assistant directly within a live hospital EHR system showing that effective clinical NLP not only depends on model quality but also on security, governance, and workflow integration. \citet{blavskovic2025robust} similarly analyzed the trade-offs between local and hosted LLMs, arguing that on-premise models can improve privacy, latency, and compliance but at the cost of greater computational and operational burden. Although recent work has begun to address privacy-preserving deployment in clinical NLP, there is still limited work on how grounded EHR QA methods compare when all training and inference must remain on commodity hardware.

\label{sec:related_work}

\section{Methods}
\label{sec:methods}

\subsection{Task Formulation}

The shared task focuses on evidence-grounded question answering over electronic health records (EHRs). 
Given a question $q$ and a clinical document $D$ consisting of sentences $\{s_1, \dots, s_n\}$, the system must identify textual evidence supporting the answer and generate a concise response grounded in that evidence. 
The task is split into four subtasks: question interpretation (Subtask 1), evidence identification (Subtask 2), answer generation (Subtask 3), and evidence alignment (Subtask 4).

\paragraph{Question Interpretation}
Patient-authored questions are often verbose and unstructured, intertwining complex personal narratives with medical queries. 
This subtask requires models to transform the raw patient narrative $q$ into a concise, clinician-interpreted question $q_{clin}$, restricted to 15 words. 
The objective is to distill the core clinical information needed into a targeted query that a clinician would write.

\paragraph{Evidence Identification}
Clinical notes provide dense, multi-faceted context spanning various events and diagnoses.
Given the question ($q$ or $q_{clin}$) and the segmented clinical document $D = \{s_1, \dots, s_n\}$, systems must extract a minimal and sufficient evidence subset $E \subseteq D$ necessary to formulate an answer.

\paragraph{Answer Generation}
This subtask challenges models to synthesize a coherent, patient-friendly answer $A$ consisting of generated sentences $\{a_1, \dots, a_m\}$, restricted to a maximum length of 75 words.
The generated response $A$ must directly address the query $q$ while remaining strictly grounded in the clinical document $D$.

\paragraph{Evidence Alignment}
The final subtask enforces explicit traceability by aligning the generated text $A$ back to the source $D$. 
Formulated as a many-to-many mapping problem, models must link each answer sentence $a_i \in A$ to a specific set of supporting evidence sentences (citations) $C_{a_i} \subseteq D$.

\subsection{Architecture Design}

\paragraph{Question Interpretation}

We use three models for this subtask: Qwen3-4B \cite{qwen3technicalreport}, Qwen2.5-14B \cite{qwen2.5}, and gpt-oss-120b
\cite{openai2025gptoss120bgptoss20bmodel}.

We evaluate different few-shot prompting setups using the first three or five cases from the development set. In particular, we test two 3-shot prompts for Qwen3-4B, three 3-shot prompts for Qwen2.5-14B, and for gpt-oss-120b we consider three 3-shot prompts and four 5-shot prompts.

Inspecting the outputs of Qwen2.5-14B shows that some answers far exceed the 15 word limit.
Hence, we conduct experiments with a two-step approach where the initial answer is provided by a 3-shot prompt using Qwen2.5-14B, and is revised by Qwen3-4B to obtain the final answer.

Finally, following \citet{leviathan2025prompt}, we also experiment with query repetition using gpt-oss-120b with the best-performing prompt on the development set.

\paragraph{Evidence Identification}

To identify sentences in the clinical note relevant to answering a question, we explore both retrieval-based approaches and supervised classification methods.

To address the limited size of the development set for fine-tuning encoder models, we generate a supplementary synthetic dataset. Using a local deployment of Llama3.1-70B \cite{llama3}, we synthesize 10 novel cases for each original development case, resulting in 200 new synthetic cases comprising 1,818 sentences annotated with three-class relevance labels. 
This generation process utilizes a two-stage pipeline consisting of an initial synthesis followed by targeted LLM-based repairs, strictly guided by manually defined quality thresholds (e.g., bounding sentence length to 10-500 characters and restricting the ratios of essential and supplementary relevance labels). Details are reported in Appendix~\ref{appendix:synth}.

Utilizing this synthetic dataset, we train a cross-encoder evidence classifier by fine-tuning a BERT-style encoder.
To capture nuanced relevance, the architecture features a shared representation layer that feeds into two distinct task heads at once: a 3-way fine-grained classification head and a 2-way binary classification head, following the multi-head training approach of HYDRA \cite{hydra}.
To prevent the synthetic cases from dominating the training signal, we balance the dataset by upsampling the real cases and downsampling the synthetic data to achieve a strict 1:1 ratio. The optimal configuration is robustly validated using case-level K-fold cross-validation.

\begin{figure*}[t]
\begin{center}
\includegraphics[width=0.9\linewidth]{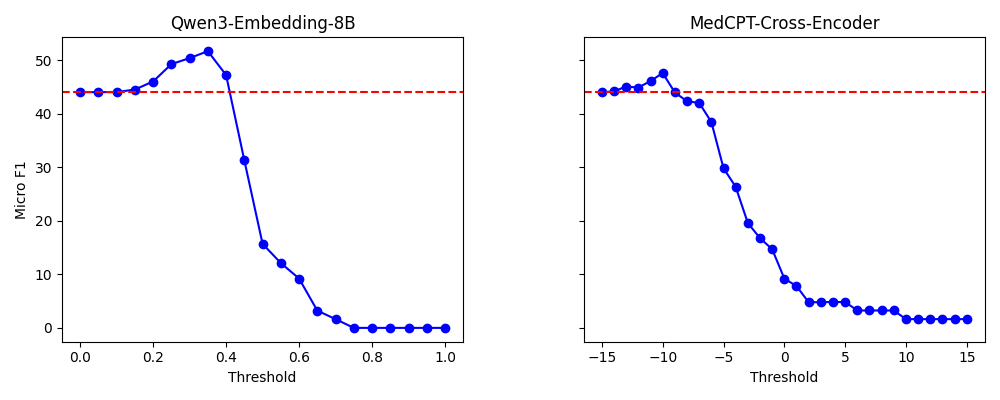}
\caption{Overall scores for different threshold values for both methods on the development set for Subtask 2. Dashed red line shows the naive baseline of classifying everything as relevant.}
\label{fig:subtask2-threshold}
\end{center}
\end{figure*}

Regarding retrieval-based approaches, we use two models with different architectures and parameter counts, namely Qwen3-Embedding-8B \cite{qwen3embedding}, and MedCPT-Cross-Encoder \cite{jin2023medcpt}, to obtain similarity scores between $q_{clin}$ and each $s_i \in D$. Then, we determine decision thresholds $t$ based on strict micro F1 scores on the development set for different $t$ values, defining the set of sentences \mbox{$E = \{\, s_i \in D \mid \mathrm{score}(q_{\text{clin}}, s_i) > t \,\}$} as relevant. See Figure \ref{fig:subtask2-threshold} for the relationship between different threshold values and F1 score.

MedCPT-Cross-Encoder provides pair-wise scores for a query ($q_{clin}$) and each $ s_i \in D$, and these scores are used directly. For Qwen3-Embedding-8B embedding model, the scores are computed as the cosine similarity between the embeddings of $q_{clin}$ and each $s_i \in D$.

\paragraph{Answer Generation}

We experiment with two setups for this subtask: a 0-shot prompt that includes the patient-authored question $q$, the clinician-interpreted question $q_{clin}$, and the clinical note excerpt $D$ with gpt-oss-120b; and a two-step 0-shot approach using $q$ and $D$ with Qwen3-4B.

\paragraph{Evidence Alignment}

To identify the supporting evidence for each answer sentence, we explore both out-of-the-box approaches using embedding and generative models, as well as supervised classification methods.

For fine-tuning our alignment cross-encoder, we use a BERT-style architecture to model the pairwise relationships among the queries ($q$ and $q_{clin}$), generated answer $A$, and source evidence $D$. 
Specifically, this classification is performed between each  $a_i \in A$ and $s_j \in D$. 
In contrast to the multi-head setup used in Subtask 2, we employ a standard binary classification head to output a relevance decision, indicating whether the specific evidence sentence supports the answer sentence, since the task does not allow for a finer-grained distinction. 
In addition, we experiment with expanding the training corpus by reusing the synthetic data generated for evidence identification (Subtask 2) as pseudo-alignments, maintaining the same 1:1 ratio between real and synthetic data to preserve training balance.

For the out-of-the-box models, we consider four approaches.
First, we adopt a threshold approach similar to Subtask 2 using Qwen3-Embedding-8B, computing similarities between each $s_i \in D$ and $a_j \in A$.
Second, we apply two-step prompting for list-wise alignment using Qwen3-4B, where $q$, $q_{clin}$, $D$, and $A$ are provided in the prompt. The first step uses a 1-shot prompt to generate a natural language response, and the second step reformats the output into JSON.
Third, we perform pair-wise alignment using a 0-shot prompt as a binary classifier with Qwen3.5-35B-A3B~\cite{qwen3.5}. For each $a_i \in A$ and $s_j \in D$, the model determines whether $s_j$ supports $a_i$.
Fourth, we use list-wise alignment with a 1-shot prompt and Qwen3.5-35B-A3B, similar to the second approach but without the reformatting step.

\begin{figure}[t]
\begin{center}
\includegraphics[width=0.85\columnwidth]{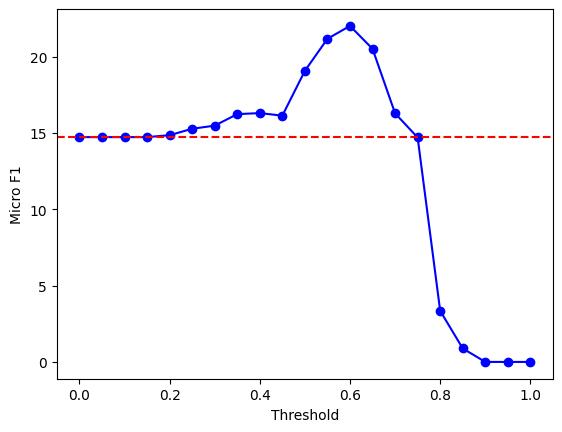}
\caption{Overall scores for different threshold values on the development set for Subtask 4. Dashed red line shows the naive baseline of classifying everything as relevant.}
\label{fig:subtask4-threshold}
\end{center}
\end{figure}

For the 1-shot prompts, the first case from the development set is used as the example. See Figure \ref{fig:subtask4-threshold} for the relationship between threshold values and F1 for the Qwen3-Embedding-8B approach.

\section{Experimental Apparatus}
\label{sec:experimentalapparatus}

\subsection{Dataset}
\label{sec:dataset}

We conduct all experiments using the ArchEHR-QA 2026 dataset \cite{soni-demner-fushman-2026-dataset}, designed to benchmark question answering over electronic health records (EHRs). 
The dataset models an interaction scenario in which patients ask questions about their health records and clinicians provide answers with explicit evidence grounding.

Each instance in the dataset, referred to as a \textit{case}, combines a patient question with a clinical note excerpt from the MIMIC database~\cite{johnson2016mimic}. Alongside the raw text, the dataset provides multiple layers of expert annotations supporting different stages of the QA pipeline.

Every case includes a free-form \textit{patient question}, a shorter \textit{clinician-interpreted question} (reference for Subtask~1), a \textit{clinical note excerpt} segmented into numbered sentences to enable sentence-level grounding (reference for Subtask~2), a \textit{clinician-written answer} addressing the question (reference for Subtask~3), \textit{evidence links} connecting answer sentences to supporting sentences from the clinical note excerpt (reference for Subtask~4), and the \textit{clinical specialty} associated with the case.

The dataset is divided into a development set with 20 cases and two test sets, \textit{test} and \textit{test-2026}, comprising 100 and 47 cases, respectively.
The labels for the test sets are withheld, and a maximum of three submissions can be evaluated.
The \textit{test-2026} set is used for Subtasks 1, 2, and 3, while Subtask 4 is evaluated on both test sets.
Dataset statistics are summarized in Table~\ref{tab:dataset_stats}.

\begin{table*}[t]
\centering
\small
\begin{tabular}{lccc}
\toprule
Statistic & Development & Test & Test-2026\\
\midrule
Cases & 20 & 100 & 47\\
Note sentences & 428 & 2599 & 1596\\
Average note sentences per case & 21.40 & 25.99 & 33.96\\
Average note sentence length  & 15.00 & 14.64 & 14.64\\
Average note length per case & 320.90 & 380.42 & 497.26\\
Average patient question length  & 16.20 & 14.64 & 14.72\\
Average clinician question length  & 10.75 & 10.63 & 10.43\\
Average answer sentences per case & 4.75 & 4.70 & 4.40\\
Average answer sentence length & 15.49 & 15.40 & 16.51\\
\bottomrule
\end{tabular}
\caption{Descriptive statistics of the ArchEHR-QA dataset across the development and two test splits. All text length measurements are reported as word counts.}
\label{tab:dataset_stats}
\end{table*}

\subsection{Models}
\label{sec:models}
To address the diverse requirements of the four ArchEHR-QA subtasks while adhering to our local hardware constraints, we evaluate a broad spectrum of parameter-efficient model architectures.
An overview of all utilized models and their respective parameter counts is provided in Table~\ref{tab:parameter_count}.

\paragraph{Autoregressive Generative Models}

For tasks requiring text generation, we experiment with several large and small instruction-tuned autoregressive language models (i.e., Qwen3-4B-Instruct~\cite{qwen3technicalreport}, Qwen2.5-14B-Instruct~\cite{qwen2.5}, Qwen3.5-35B~\cite{qwen3.5}, and gpt-oss-120b~\cite{openai2025gptoss120bgptoss20bmodel}).
To enable local execution of larger models, we run quantized versions converted to the MLX~framework~\cite{mlx2023} with 4-bit precision.

\paragraph{Decoder-only Embedding Models}
For similarity-based retrieval and alignment tasks, we utilize the Qwen3-Embedding-8B~\cite{qwen3embedding} model. 
This model produces dense vector representations, enabling efficient semantic similarity computation using cosine similarity. 
Additionally, we employ the MedCPT cross-encoder~\cite{jin2023medcpt}, a domain-specific model trained on biomedical literature, to compute direct pairwise relevance scores between queries and clinical text.

\paragraph{Encoder-only Classification Models}
For binary and multi-class sentence classification, we rely on parameter-efficient, BERT-style encoder architectures.
These models are well-suited for local fine-tuning and rapid cross-encoder inference and are known to be strong short-text classifiers~\cite{short-text}.
We evaluate domain-specific models and strong general-domain encoders.

\begin{table*}[t]
\centering
\small
\begin{tabular}{ll}
\toprule
Model & \#Parameter \\
\midrule
\textit{Autoregessive Generative Models}\\
mlx-community/gpt-oss-120b-MXFP4-Q4 & 117B total / 5.1B active \\
RepublicOfKorokke/Qwen3.5-35B-A3B-mlx-lm-mxfp4 & 35B total / 3B active \\
Qwen/Qwen2.5-14B-Instruct & 14.7B \\
Qwen/Qwen3-4B-Instruct-2507 & 4B \\
\midrule
\textit{Retrieval Models}\\
Qwen/Qwen3-Embedding-8B & 8B \\
ncbi/MedCPT-Cross-Encoder & 109M \\ %
\midrule
\textit{Classification Models}\\
microsoft/deberta-base & 138M \\ %
emilyalsentzer/Bio\_ClinicalBERT & 108M \\ %
\bottomrule
\end{tabular}
\caption{Evaluated models categorized by architecture type and sorted by decreasing parameter count. All models are obtained from Hugging Face (B: billions, M: millions).}
\label{tab:parameter_count}
\end{table*}

\subsection{Hyperparameter Optimization}
\label{sec:hyperparameteroptimization}

For all generative inference-only experiments, we use the default decoding parameters, with a temperature of 0.7 and a top‑p value of 0.9.
No additional tuning is performed for these settings.

For the fine-tuning of our classifier in Subtask 2: Evidence Identification and Subtask 4: Evidence Alignment, we conduct a grid search to identify optimal hyperparameters.
The hyperparameter search space includes learning rates in $\{1\times10^{-5}, 2\times10^{-5}, 3\times10^{-5}, 5\times10^{-5}\}$, training epochs in $\{1, 2\}$, and dropout probabilities in $\{0.10, 0.20\}$. 
The optimal parameters are determined by evaluating performance across 5-fold cross-validation runs on the development set. 
The selected hyperparameters are reported in Table~\ref{tab:final_hyparams}.

\begin{table*}[t]
\centering
\small
\begin{tabular}{lllll}
\toprule
Task & Model & LR & Epochs & Dropout \\
\midrule
Subtask 2 & Bio\_ClinicalBERT      & $2\times10^{-5}$ & 1 & 0.1 \\
Subtask 2 & DeBERTa-base           & $2\times10^{-5}$ & 1 & 0.1 \\
\midrule
Subtask 4 & Bio\_ClinicalBERT      & $3\times10^{-5}$ & 1 & 0.1 \\
Subtask 4 & DeBERTa-base           & $2\times10^{-5}$ & 1 & 0.1 \\
\bottomrule
\end{tabular}
\caption{Optimized hyperparameters for the BERT-style classifiers in Subtask 2 and Subtask 4. Values are selected via a grid search evaluated with 5-fold cross-validation on the development set.}
\label{tab:final_hyparams}
\end{table*}

\subsection{Metrics}
\label{sec:metrics}

To evaluate system performance across the four subtasks, we employ a combination of established classification and text-generation metrics.

For the generative tasks (Subtasks 1 and 3), we report several evaluation metrics to capture various aspects of text quality. 
We use‚ BLEU \cite{papineni2002bleu}, ROUGE \cite{lin2004rouge} %
SARI \cite{xu2016optimizing}, AlignScore \cite{zha2023alignscore}, and MEDCON \cite{yim2023aci}.
Beyond lexical overlap, we employ BERTScore \cite{bert-score} to compute the semantic similarity between the generated and reference sentences using contextual embeddings.

For the extraction and alignment tasks (Subtasks 2 and 4), performance is measured using standard micro Precision, micro Recall, and micro F1 scores. 

We strictly followed the shared task's official evaluation scripts for all subtasks. For Subtask 4, note that the script computes metrics in a specific manner that may appear counterintuitive (e.g., not reaching 100\% recall in the all-relevant configurations), as defined by the task organizers.

\subsection{Hardware}
\label{sec:hardware}

A core motivation of our work is to demonstrate the feasibility of executing EHR question-answering pipelines entirely on commodity hardware.
All experiments are therefore conducted locally on Apple Silicon devices rather than GPU clusters.

Most‚ of our experiments are performed on an Apple MacBook M4 Pro with 48GB of unified memory.
This system is sufficient for training the classifier models and running most inference experiments.

To further evaluate the feasibility of running larger models locally, we conduct experiments on high-end consumer hardware. 
Specifically, we use an Apple Mac Studio M3 Max with 96GB of memory to run experiments with the gpt-oss-120B model.

\section{Results}
\label{sec:results}

\paragraph{Question Interpretation}

We select the three best-performing setups on the development set, based on BERTScore, for evaluation on the test set. Namely, gpt-oss-120b with a 5-shot prompt, the repeated-query approach with the same prompt, and the two-step approach using smaller models. Table~\ref{tab:subtask1_results} shows the development and test set results for these models. Overall score corresponds to the average of the four scores. See Appendix~\ref{appendix:prompts} for the prompt template corresponding to the single-query setup, which achieved the best overall score.

\begin{table}[t]
\centering
\small
\begin{tabular}{lccc}
\toprule
Metric & SQ & DQ & TS \\
\midrule

\textit{Dev Set}\\
ROUGE-Lsum & 23.51 & 24.24 & \textbf{28.74} \\
BERTScore  & 41.15 & \textbf{43.79} & 41.94 \\
\midrule
\textit{Test Set}\\
ROUGE-Lsum & 22.92 & \textbf{23.09} & 21.19 \\
BERTScore & 36.93 & \textbf{37.29} & 33.10 \\
AlignScore & 21.42 & 20.22 & \textbf{22.91} \\
MEDCON (UMLS) & 21.26 & \textbf{21.82} & 21.18 \\
\textit{Overall Score} & \textbf{25.63} & 25.61 & 24.59 \\
\bottomrule
\end{tabular}
\caption{Test and development set results for Subtask 1. Only ROUGE and BERTScore are used on the development set. SQ: Single Query, DQ: Double Query, TS: Two Step.}
\label{tab:subtask1_results}
\end{table}

\paragraph{Evidence Identification}

For this subtask, we compare our methods to the naive baseline of considering $E = D$, that is considering every sentence as relevant evidence on the development set. Surprisingly, most methods struggle to outperform this baseline substantially. See Table~\ref{tab:strict_results} for strict micro precision, recall, and F1 scores for every method on the development set. Based on these results, we selected Qwen3-Embedding-8B, and Bio\_ClinicalBERT-synth-HYDRA approaches to be evaluated on the test set, where they had overall scores of 51.61 and 44.43 respectively, which can also be seen on the same table.

\begin{table*}[tb]
\centering
\small
\begin{tabular}{lcccccc}
\toprule
 & \multicolumn{3}{c}{Dev Set} & \multicolumn{3}{c}{Test Set} \\
\cmidrule(lr){2-4} \cmidrule(lr){5-7}
Setup & Strict P & Strict R & Strict F1 & Strict P & Strict R & Strict F1 \\
\midrule

\textit{Baseline}\\
Everything is relevant             & 28.27 & 100.00 & \underline{44.08}  & -- & -- & -- \\

\midrule
\textit{Similarity Threshold}\\
Qwen3-Embedding-8B                 & 48.55 & 55.37 & \underline{\textbf{51.74}} & 37.92 & 80.74 & \textbf{51.61}\\
MedCPT-Cross-Encoder               & 34.04 & 79.34 & 47.64  & -- & -- & --\\

\midrule
\textit{5-Fold Training }\\
Bio\_ClinicalBERT-real-only-binary-head      & 31.39 & 93.39 & 46.99 & -- & -- & -- \\
Bio\_ClinicalBERT-real-HYDRA                 & 32.46 & 92.56 & 48.07 & -- & -- & -- \\
Bio\_ClinicalBERT-synth-HYDRA                & 36.01 & 85.12 & \underline{50.61} & 32.40 & 70.68 & 44.43 \\
DeBERTa-real-only-binary-head      & 28.72 & 94.21 & 44.02 & -- & -- & -- \\
DeBERTa-real-HYDRA                 & 28.27 & 100.00 & 44.08 & -- & -- & -- \\
DeBERTa-synth-HYDRA                & 28.97 & 93.39 & 44.23 & -- & -- & -- \\

\bottomrule
\end{tabular}
\caption{Subtask 2 results. P and R stand for Precision and Recall, respectively. ``Synth'' indicates training with synthetic data, while ``HYDRA'' denotes multiple heads. Development set results for fine-tuned models are computed from out-of-fold predictions. \textbf{Bold} values mark the highest overall F1 score, and \underline{underlined} values show the best F1 within each method category on the development set.
}
\label{tab:strict_results}
\end{table*}

\paragraph{Answer Generation}

Table~\ref{tab:subtask3_results} shows development results for both approaches and test results for the approach with the best performance (gpt-oss-120b). The prompt template is included in Appendix~\ref{appendix:prompts}.

\begin{table}[t]
\centering
\small
\begin{tabular}{lccc}
\toprule
Metric & \multicolumn{2}{c}{Dev Set} & Test Set \\
\cmidrule(lr){2-3} \cmidrule(lr){4-4}
       & Qwen & gpt & gpt \\
\midrule
Overall Score & 27.76 & 32.38 & 31.51 \\
\midrule
BLEU          & 1.19 & 4.34 & 3.82 \\
ROUGE-Lsum         & 15.73 & 21.34 & 22.04 \\
SARI          & 49.75 & 58.28 & 56.55 \\
BERTScore     & 30.95 & 39.66 & 39.30 \\
AlignScore    & 41.16 & 38.29 & 33.92 \\
MEDCON (UMLS) & -- & -- & 33.44 \\
\bottomrule
\end{tabular}
\caption{Evaluation results for Subtask 3 on the development and test sets. For the development set MEDCON score was not calculated and therefore was also not included in the overall score, which is the mean of all other scores.}
\label{tab:subtask3_results}
\end{table}

\paragraph{Evidence Alignment}

Similar to Evidence Identification, we again compare our methods to the naive baseline of considering $C_{a_i} = D$, $\forall a_i \in A$; that is considering every excerpt sentence as supporting evidence for every answer sentence, on the development set. See Table~\ref{tab:subtask4-kfold-mean-std} for strict micro precision, recall, and F1 scores for every method on the development set. Based on these results, we selected Qwen3-Embedding-8B, Qwen3.5-35B list-wise, and Bio\_ClinicalBERT approaches to be evaluated on the test set, where they had overall scores of 59.45, 74.84, and 8.33 respectively, which can also be seen on the same table. See Appendix~\ref{appendix:prompts} for the prompt template for the setup with the best overall score (Qwen3.5-35B list-wise).

\begin{table*}[tb]
\centering
\small
\begin{tabular}{lcccccc}
\toprule
 & \multicolumn{3}{c}{Dev Set} & \multicolumn{3}{c}{Test Set} \\
\cmidrule(lr){2-4} \cmidrule(lr){5-7}
Model & Micro-P & Micro-R & Micro-F1 & Micro-P & Micro-R & Micro-F1 \\
\midrule

\textit{Baseline}\\
Everything is relevant             & 8.16 & 75.11 & \underline{14.73} & -- & -- & -- \\

\midrule
\textit{Similarity Threshold}\\
Qwen3-Embedding-8B                 & 22.82 & 21.27 & \underline{22.01} & 52.41 & 68.69 & 59.45 \\

\midrule
\textit{Prompting}\\
Qwen3-4B             & 20.00 & 11.76 & 14.81 & -- & -- & --  \\
Qwen3.5-35B pair-wise              & 35.35 & 15.84 & 21.88 & -- & -- & -- \\
Qwen3.5-35B list-wise              & 31.58 & 21.72 & \underline{\textbf{25.74}} & 79.65 & 70.57 & \textbf{74.84} \\

\midrule
\textit{5-Fold Training}\\
Bio\_ClinicalBERT                  & 19.14 & 24.43 & \underline{21.47} & 6.26 & 12.46 & 8.33 \\
Bio\_ClinicalBERT + Synth          & 16.27 & 25.33 & 19.82 & -- & -- & -- \\
DeBERTa-base                       & 9.24 & 50.67 & 15.63 & -- & -- & --  \\
DeBERTa-base + Synth               & 12.67 & 45.25 & 19.80 & -- & -- & --  \\

\bottomrule
\end{tabular}
\caption{Subtask 4 results. Micro-Precision (P), Micro-Recall (R), and Micro-F1 are reported. ``Synth'' denotes training augmented with synthetic data. Development set results for fine-tuned models are derived from merged out-of-fold predictions.
\textbf{Bold} values denote the highest overall F1 score, and \underline{underlined} values indicate the best F1 score within each method category on the development set.}
\label{tab:subtask4-kfold-mean-std}
\end{table*}

\section{Discussion}
\label{sec:discussion}

Results from Subtasks 1 and 3 indicate that although larger models generally achieve stronger performance, the performance gap compared to smaller models remains moderate. Comparable results can therefore be obtained with much smaller architectures, suggesting that model scale alone is not the main driver of performance for these tasks.

For Subtask 1, we evaluated prompt repetition as proposed by \citet{leviathan2025prompt}. In contrast to their findings, in our case the prompt repetition did not considerably outperform the single-query approach. Nevertheless, it produced slightly higher results on three out of the four evaluation metrics on the test set, indicating a modest but consistent improvement across most metrics.

The results for Subtask 2 reveal a strong recall bias across many approaches. Notably, the baseline that classifies every candidate as relevant achieves perfect recall and surprisingly competitive F1 performance. This observation suggests that the dataset implicitly rewards high recall and that reliably identifying truly irrelevant contexts remains challenging. More advanced methods, such as embedding similarity or cross-encoder ranking, generally increase precision but often at the cost of reduced recall. Interestingly, models trained with HYDRA and synthetic data exhibit somewhat more balanced precision–recall trade-offs in certain cases. This may indicate that these models benefit from additional training signals and that synthetic examples can help establish clearer decision boundaries when real training data is limited.

A comparison between DeBERTa and Bio\_ClinicalBERT further highlights the importance of domain-specific modeling. The biomedical model consistently performs better in Subtask 2 than the general-purpose model, suggesting that the task contains considerable domain-specific terminology and contextual cues that are better captured by specialized pretraining.

Unlike the other subtasks, Subtask 4 exhibits a substantial discrepancy between the development and test set performance. We hypothesize that this may be caused by a distribution shift between the two splits. This could also explain the severe degradation observed for the fine-tuned model on the test set compared to out-of-the-box models. Since the fine-tuned models were trained on the full development set using hyperparameters derived from k-fold validation, they may have overfitted to the development distribution, which does not fully reflect the characteristics of the test set.

Similar to Subtask 2, the biomedical model again outperforms the general-domain model in Subtask 4. However, unlike Subtask 2, synthetic data improves performance for the general-domain model (DeBERTa) but not for the biomedical model (Bio\_ClinicalBERT). Considering the more balanced precision–recall behavior observed in Subtask 4, this suggests that synthetic data may be particularly beneficial when model predictions are highly skewed, as is the case in Subtask 2.

\subsection{Future Work and Impact}
\label{sec:futurework}

Our findings establish a solid baseline for local EHR QA.
However, our setups still rely on high-end commodity hardware. 
Future work should explore the feasibility of even smaller, highly optimized language models capable of executing on highly constrained edge devices, such as mobile phones or clinical tablets, to further lower the barrier to entry.

Furthermore, this shared task operates under low-resource data, which likely contributes to the generalization challenges observed with our fine-tuned cross-encoders.
To mitigate this data scarcity, future research should explore integrating large-scale proxy datasets from related clinical domains alongside more advanced synthetic data generation methods to better stabilize model fine-tuning and improve robustness to distribution shifts.

\section{Conclusion}
\label{sec:conclusion}

As shown on the competition leaderboards, our approaches implemented using only commodity hardware, achieve competitive results across multiple subtasks. For Subtask~1\footnote{\href{https://www.codabench.org/competitions/12865/}{www.codabench.org/competitions/12865/}}
 and Subtask~4\footnote{\href{https://www.codabench.org/competitions/13528/}{www.codabench.org/competitions/13528/}}
, our submissions perform above the average score. Although our results for Subtask~3\footnote{\href{https://www.codabench.org/competitions/13527/}{www.codabench.org/competitions/13527/}}
 are lower, they remain comparable to those of other systems. For Subtask~2\footnote{\href{https://www.codabench.org/competitions/13526/}{www.codabench.org/competitions/13526/}}, our approach performs less favorably, indicating substantial room for improvement.

Our experiments further demonstrate that, with appropriate configurations and training strategies, smaller models can achieve performance comparable to that of significantly larger models. This highlights the potential of resource-efficient approaches for tasks in the clinical domain.

Overall, our findings suggest that privacy-preserving, fully local electronic health record (EHR) question-answering systems are technically feasible using current models and accessible hardware. While such systems may still lag behind large-scale cloud-based solutions in certain aspects, the performance gap appears small enough for many real-world applications in which data privacy and on-device processing are critical requirements.

\section*{Limitations}
While our approach demonstrates the feasibility of local EHR QA, several limitations remain.
Although our setup avoids external cloud infrastructure, it still relies on high-end commodity hardware (e.g., an Apple MacBook with an M4 Pro chip and 48 GB memory), which may not be readily available in all under-resourced clinical environments. 
Furthermore, our evaluation is constrained to the ArchEHR-QA dataset, limiting the immediate generalizability of our findings across diverse clinical specialties or multilingual patient populations.

\section*{Ethical Considerations}
Our work is primarily motivated by the ethical imperative to protect patient privacy. 
By executing all models entirely on a local notebook, we eliminate the need to transmit sensitive Protected Health Information to external, third-party cloud APIs. 
However, the use of generative AI in healthcare carries inherent risks for patient safety. 
The models we evaluate are prone to hallucinations and false conclusions.
In a clinical context, such errors could negatively impact patient care if accepted without careful review.

\section{Bibliographical References}\label{sec:reference}

\bibliographystyle{lrec2026-natbib}
\bibliography{lrec2026-example}

\appendix

\section{Prompt Templates}
\label{appendix:prompts}

\begin{lstlisting}[breaklines=true, basicstyle=\small\ttfamily, caption={Best-performing prompt template for Subtask 1: Question Interpretation}, label={lst:question-int}]
You rewrite long, messy questions into ONE short core question, phrased in third person from a clinician's perspective.

Task:
- Identify the single core question being asked.
- Rewrite it as one clear question in THIRD PERSON.

Hard Rules (must follow):
- Output exactly ONE sentence
- Maximum 15 words
- Use third-person only (e.g., the patient, he, she, they)
- Do NOT explain, summarize, or add context
- Do NOT add new information
- Output ONLY the rewritten question

Question Style:
- Use simple forms like ``What'', ``Why'', ``Did'', ``Is'', ``How''

Examples:

<5 Examples>
\end{lstlisting}

\begin{lstlisting}[breaklines=true, basicstyle=\small\ttfamily, caption={Best-performing prompt template for Subtask 3: Answer Generation}, label={lst:answer-gen}]
You are a clinical documentation assistant.

Your task is to generate a professional, natural-language answer to the patient's question using ONLY the information explicitly stated in the provided clinical note excerpt.

Instructions:
- Use only facts supported by the clinical note excerpt.
- Do NOT add outside medical knowledge.
- Do NOT speculate or infer beyond what is documented.
- If the note does not fully answer the question, provide a faithful response limited to the documented information.
- Write in a professional clinical tone.
- Limit your response to a maximum of 75 words (approximately 5 sentences).
- Do not include citations or sentence numbers.
- Do not mention the instructions in your response.
\end{lstlisting}

\begin{lstlisting}[breaklines=true, basicstyle=\small\ttfamily, caption={Best-performing prompt template for Subtask 4: Evidence Alignment}, label={lst:evidence-alignment}]
You are a clinical evidence alignment system.

Your task is to align each answer sentence to supporting clinical note sentences.

Important rules:

1. Alignment is at the answer sentence level.
2. Only include note sentences that DIRECTLY support the answer sentence.
3. Do NOT over-cite.
4. Do NOT under-cite.
5. Do NOT infer beyond the note text.
6. Use only the provided numbered note sentences.
7. If no supporting evidence exists, return an empty list [].
8. Evidence IDs must be strings.
9. Output must be STRICT JSON.
10. Do not include explanations or reasoning.

Output format:

[
  {
    "case\_id": "<case\_id>",
    "prediction": [
      {
        "answer\_id": "<ID>",
        "evidence\_id": ["<note\_id>", ...]
      }
    ]
  }
]

Example:

<1 Example>
\end{lstlisting}

\section{Synthetic Data Generation Details}
\label{appendix:synth}

To augment the limited development set for fine-tuning our encoder models, we generate synthetic clinical cases using a local deployment of Llama-70B via Ollama.
For each of the 20 original development cases, we generate 10 novel synthetic variations, yielding a total of 200 new cases. 
This process produces 1,818 synthetic sentences, each assigned a 3-class relevance label (``essential'', ``supplementary'', or ``not-relevant'').

The generation is executed in a two-stage pipeline: an initial LLM generation phase followed by a targeted LLM-based repair phase to correct formatting and distribution errors.
Both steps are strictly guided by the following manually defined quality thresholds to ensure the synthetic data closely mirrors the distribution of the real dataset:

\begin{itemize}
    \item \textbf{Sentence Count constraints:} Between 10 and 20 sentences per case.
    \item \textbf{Sentence Length constraints:} Between 10 and 500 characters per sentence.
    \item \textbf{Relevance Label Ratios:}
    \begin{itemize}
        \item Essential ratio: $0.10 \leq r_{ess} \leq 0.40$
        \item Supplementary ratio: $r_{sup} \leq 0.15$
        \item Not-Relevant ratio: $r_{nr} \geq 0.45$
    \end{itemize}
\end{itemize}

The following prompt template is used during the initial generation phase to construct the synthetic cases based on the provided seed examples from the development set.

\begin{lstlisting}[breaklines=true, basicstyle=\small\ttfamily, caption={Prompt template for generating synthetic clinical QA cases.}, label={lst:synthetic_prompt}]
You are a medical data augmentation assistant. Your task is to create a synthetic clinical note excerpt with relevance labels for training a medical evidence identification model.

TASK: Study the real examples below, then generate a NEW synthetic case with DIFFERENT medical content (different condition, treatment, or scenario). Pay close attention to the label distributions in the examples - most sentences are "not-relevant".

{examples_block}

INSTRUCTIONS:
1. Create a NEW patient question about a DIFFERENT medical topic from any of the examples above.
2. Create a corresponding clinician question (concise, max 15 words).
3. Write a LONG clinical note excerpt with 15-30 sentences. Real clinical notes are long and contain many sections (History of Present Illness, Hospital Course, Medications, Labs, Vitals, Plan, etc.). Include realistic clinical detail.
4. Label each sentence as "essential", "supplementary", or "not-relevant":
  - essential: Directly answers the patient's question (~20-35%
  - supplementary: Provides helpful context but not critical (~5-10%
  - not-relevant: Does NOT help answer the question (~55-70%

IMPORTANT DISTRIBUTION GUIDANCE:
- Most sentences in a clinical note are NOT relevant to the patient's specific question. The note covers many topics (vitals, labs, medications, procedures, consults, discharge planning) and only a fraction directly address the question.
- "supplementary" is the RAREST label - only a few sentences at most.
- "not-relevant" should be the MOST COMMON label - typically 60%
- Look at the examples above: notice how the majority of sentences are labeled "not-relevant".

OUTPUT FORMAT (use exactly this JSON format):
{
  "patient_question": "...",
  "clinician_question": "...",
  "sentences": [
    { "id": "1", "text": "..." },
    { "id": "2", "text": "..." }
  ],
  "relevance_labels": [
    { "sentence_id": "1", "relevance": "essential" },
    { "sentence_id": "2", "relevance": "not-relevant" }
  ]
}

Generate a synthetic case now. Output ONLY valid JSON, no explanation.
\end{lstlisting}

\end{document}